\documentclass[10pt,onecolumn,letterpaper]{article}

\usepackage{cvpr}
\usepackage{times}
\usepackage{epsfig}
\usepackage{graphicx}
\usepackage{amsmath}
\usepackage{amssymb}
\usepackage{verbatimbox}
\usepackage{algorithm} 
\usepackage{algpseudocode} 
\usepackage[normalem]{ulem}
\usepackage{amsmath}
\usepackage{cite}
\newcommand{\stkout}[1]{\ifmmode\text{\sout{\ensuremath{#1}}}\else\sout{#1}\fi}
\algnewcommand\And{\textbf{and}}
\usepackage{array}
\usepackage{multirow}

\usepackage[breaklinks=true,
            bookmarks=false,
            colorlinks = true,
            linkcolor = black,
            urlcolor  = black,
            citecolor = black,
            anchorcolor = black]{hyperref}

\cvprfinalcopy 

\def\cvprPaperID{****} 

\begin{document}

\title{Deep Compression for PyTorch Model Deployment on Microcontrollers}

\author{Eren Dogan\\
Ozyegin University\\
Istanbul, 34794, Turkey\\
{\tt\small eren.dogan@ozu.edu.tr}
\and
H. Fatih Ugurdag\\
Ozyegin University\\
Istanbul, 34794, Turkey\\
{\tt\small fatih.ugurdag@ozyegin.edu.tr}
\and
Hasan Unlu\\
Stanford University\\
Stanford, CA 94305, USA\\
{\tt\small hasanunlu9@gmail.com}
}

\maketitle

{
\centering
\begin{abstract}
\centering\begin{minipage}{\dimexpr\paperwidth-10cm}
Neural network deployment on low-cost embedded systems, hence on microcontrollers (MCUs), has recently been attracting more attention than ever. Since MCUs have limited memory capacity as well as limited compute-speed, it is critical that we employ model compression, which reduces both memory and compute-speed requirements. In this paper, we add model compression, specifically Deep Compression~\cite{songhan}, and further optimize Unlu's work~\cite{hasanunlu}, which efficiently deploys PyTorch models on MCUs. First, we prune the weights in convolutional and fully connected layers. Secondly, the remaining weights and activations are quantized to 8-bit integers from 32-bit floating-point. Finally, forward pass functions are compressed using special data structures for sparse matrices, which store only nonzero weights (without impacting performance and accuracy). In the case of the LeNet-5 model~\cite{lecun}, the memory footprint was reduced by $12.45\times$, and the inference speed was boosted by $2.57\times$.
\end{minipage}
\end{abstract}
\par
}

\section{Introduction}
Although neural networks are trained on modern GPUs with drivers specifically developed for neural network operations, most of the use cases of these neural networks require deployment on embedded systems. Recently, Unlu~\cite{hasanunlu} proposed an efficient method for deployment of Convolutional Neural Networks (CNNs) on microcontrollers. We were able to further improve the approach in~\cite{hasanunlu} by combining it with model compression methods.
\par\medskip
In recent years, there have been many different approaches to compressing CNNs in order to increase inference speed and deploy them easily on embedded systems. One of the most impressive approaches is “Deep Compression” proposed by Han et al.~\cite{songhan}, where neural networks are pruned and quantized. In this paper, a similar approach will be followed to compress models. There are open source toolkits that implement model compression algorithms. One such example is Microsoft’s Neural Network Intelligence (NNI) toolkit~\cite{nni}, which we use in this work for pruning and quantizing the weights in convolutional (CONV) and fully connected (FC) layers of our input models.

\section{Training and Compressing the Model}
\cite{hasanunlu} trained LeNet-5 on MNIST dataset~\cite{mnist} using PyTorch library~\cite{pytorch}. We used the same model and trained it on the same dataset. We decided to apply Deep Compression~\cite{songhan} (DC) for network compression. Therefore, we did unstructured pruning on our model and implemented that using NNI’s Level Pruner to prune a specific percentage (sparsity) of weights. To find the optimal sparsity value, we did binary search.

\begin{algorithm}
    \caption{Prune the model using binary search for the optimal sparsity value\newline
    $model$: pretrained model\newline
    $tolerated\_acc\_loss$: tolerated accuracy loss of the pruned model\newline
    $min\_search\_step$: minimum step to stop the algorithm\newline
    $current\_model$: pretrained neural network model}
    \label{alg:binary}
	\begin{algorithmic}[0]
	\State $step \leftarrow 0.5$
	\State $sparsity \leftarrow 0.5$
	\State $best\_sparsity \leftarrow 0$
	\State $initial\_accuracy \leftarrow evaluate\_model(model)$
	\While {$step > min\_search\_step$}
        \State $step \leftarrow step / 2$
        \State $model \leftarrow prune\_model(current\_model, sparsity)$
        \State $accuracy \leftarrow train\_model(model)$
        \If {$accuracy \geq initial\_accuracy - tolerated\_accuracy\_loss$}
            \State $best\_model \leftarrow model$
            \State $best\_sparsity \leftarrow sparsity$
            \State $sparsity \leftarrow sparsity + step$
        \Else
            \State $sparsity \leftarrow sparsity - step$
        \EndIf
    \EndWhile
	\end{algorithmic} 
\end{algorithm}

\par\medskip
After applying Algorithm~\ref{alg:binary} to our model, we have a pruned the LeNet-5 model within tolerated accuracy bounds. According to DC, pruned models can be saved as compressed sparse format and index differences can be used instead of absolute positions of weights. Since \cite{hasanunlu} uses one dimensional arrays as weight matrices, we flattened the weight matrices and followed the same approach as DC by converting weight matrices to Compressed Sparse Column (CSC) format. Indices are saved as 8-bit unsigned integer, which is one of the smallest data types defined in the standard library of C (stdint.h) and nonzero weights are saved as 32-bit signed floating-point.
\par\medskip
DC paper~\cite{songhan} states that quantization by weight sharing can further compress the model. However, after experimenting with weight sharing, we realized that it reduces the accuracy of the model significantly, and retraining shared weights does not seem to recover this accuracy loss. Therefore, we decided to use NNI’s Naive Quantizer to quantize all weights to 8-bit integers. Since DC’s quantization also reduces bitwidth of weights to 8 for CONV layers and 5 for FC layers, our approach achieves the same compression because the smallest data type defined in the standard library of C language is 8-bit. Quantization also improved pruning performance, because some weights are quantized to zero, which we pruned later on. After quantization, all weights are saved as 8-bit signed integer, and the scale value of each layer is saved as 32-bit signed floating-point.
\par\medskip
Since quantizing activations can further improve the inference performance of the model, each layer’s output is quantized with affine quantization unlike weights, which are quantized by scale quantization. In order to prevent accuracy loss, each output array is quantized separately. Otherwise, precision loss would significantly degrade the accuracy. Maximum and minimum activation values for each output array are calculated with the training data, and these bounds are used to quantize and clip activations during inference. Scale values of outputs are stored as 32-bit signed floating-point.

\section{Improving CNN Algorithms}
Since we store weights in CSC format, we need to modify the existing algorithms of~\cite{hasanunlu} so that we can use CSC arrays directly. Otherwise, we would have needed to iterate over each CSC array numerous times, and that would reduce runtime performance significantly. The following algorithms iterate over each CSC array only once so that computations do not slow down while receiving weight values.
\par\medskip
Algorithm~\ref{alg:dot} searches for nonzero weights by iterating over the index difference array of CSC when necessary during FC layer calculations. Since it skips pruned weights, inference speed is increased significantly. Contrary to FC layers, CONV layer calculations use the same weight kernel multiple times. Therefore, Algorithm~\ref{alg:conv} receives weights kernel by kernel which are going to be multiplied with the input data later in the convolutional calculations.

\begin{algorithm}
    \caption{Perform dot product for fully connected layer using CSC arrays directly\newline
    $indices$: index difference array for CSC\newline
    $weights$: nonzero weight values as CSC\newline
    $input$: array of input values\newline
    $r, c$: row and column of weights matrix}
    \label{alg:dot}
	\begin{algorithmic}[0]
	\State $sum \leftarrow 0$
	\State $s \leftarrow 0$
	\State $sum\_indices \leftarrow indices[0]$
	\For {$i \leftarrow 0$ \textbf{to} $c$}
        \State $sum \leftarrow 0$
        \For {$j \leftarrow 0$ \textbf{to} $r$}
            \State $index \leftarrow i * r + j$
            \While {$sum\_indices < index$}
                \State $s \leftarrow s + 1$
                \State $sum\_indices \leftarrow sum\_indices + indices[s]$
            \EndWhile
            \If {$sum\_indices == index$}
                \State $sum \leftarrow sum + input[j] * weights[s]$
            \EndIf
        \EndFor
    \EndFor
	\end{algorithmic} 
\end{algorithm}

\begin{algorithm}
    \caption{Perform convolutional calculation using CSC arrays directly\newline
    $indices$: index difference array for CSC\newline
    $weights$: nonzero weight values as CSC\newline
    $input$: array of input values.\newline
    $r, c$: row and column of weights matrix\newline
    $in\_channel$: number of channels of the input\newline
    $out\_channel$: number of channels of the output}
    \label{alg:conv}
	\begin{algorithmic}[0]
	\State $s \leftarrow 0$
	\State $sum\_indices \leftarrow indices[0]$
	\State $weight\_size \leftarrow r * c * in\_channel$
	\State $weight\_kernel \leftarrow [0$ \textbf{for} $weight\_size]$
	\For {$i \leftarrow 0$ \textbf{to} $out\_channel$}
        \State $weight\_begin\_index \leftarrow i * weight\_size$
        \For {$j \leftarrow 0$ \textbf{to} $weight\_size$}
            \State $index \leftarrow weight\_begin\_index + j$
            \While {$sum\_indices < index$}
                \State $s \leftarrow s + 1$
                \State $sum\_indices \leftarrow sum\_indices + indices[s]$
            \EndWhile
            \If {$sum\_indices == index$}
                \State $weight\_kernel[j] \leftarrow weights[s]$
            \Else
                \State $weight\_kernel[j] \leftarrow 0$
            \EndIf
        \EndFor
        \State $convolution(input, weight\_kernel)$
    \EndFor
	\end{algorithmic} 
\end{algorithm}

\section{Implementation}
In this section, we discuss our implementation of the approach and compare it with~\cite{hasanunlu}'s implementation~\cite{hasanunlurepo}. Our approach described above is implemented as a C file generator. The source code of our generator is available on GitHub~\cite{repo}. The generator is written as a Python script.  Algorithm~\ref{alg:dot} and Algorithm~\ref{alg:conv} are implemented as a C header file similar to~\cite{hasanunlurepo}.
\par
\begin{figure}[t]
\begin{center}
    \includegraphics[scale=0.8]{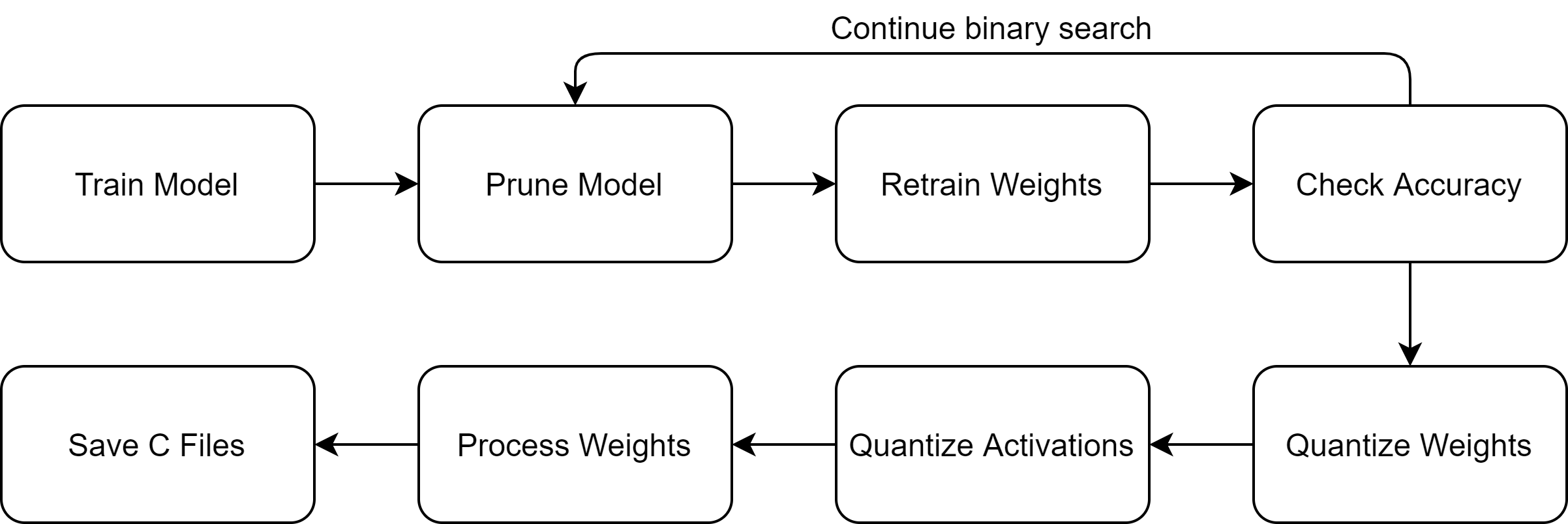}
\end{center}
    \caption{Flow of the compressed PyTorch model deployment generator.}
    \label{fig:flowchart}
\end{figure}

\medskip
The flow of the C file generator program is illustrated in Figure~\ref{fig:flowchart}. Our generator's inputs are the PyTorch model, the optimizer, and the number of epochs. The given PyTorch model does not have to be pre-trained; however, initial training can be disabled for pre-trained models to speedup the process. After pruning and quantization is completed, weights and biases are extracted from the PyTorch model and processed to be saved as CSC arrays in the output header file. Finally, main.c and main.h files are saved to be compiled and deployed on any device that can run the compiled executable file.

\par\medskip
Although our generator is a mostly automated program, parameter fine tuning might be necessary for complex and dense CNNs. For instance, some networks are more sensitive to the input data than others. In this case, input data may not be quantized for higher accuracy. Our generator has an option to decide whether input data shall be quantized.
\par\medskip
Pruning and quantization do not appear in ~\cite{hasanunlu}'s implementation. After pruning and quantization phases are completed, our implementation runs similar to~\cite{hasanunlurepo}. Weights from each layer of the PyTorch model are extracted and flattened. Unlike~\cite{hasanunlurepo} where flattened weight arrays are written on the header file, our generator processes weights by converting them to CSC arrays. After weights are processed, weights and biases are saved as a header file. Both~\cite{hasanunlurepo} and~\cite{repo} extracts weights and biases from models made with PyTorch's basic building blocks. Basic building blocks supported by these generators are Conv2d, MaxPool2d, Linear, Flatten, and ReLU.

\section{Results}
All results were obtained on an Intel Core i7-8750H processor. We trained LeNet-5 on MNIST database using the same configuration as~\cite{hasanunlu}, Adam Optimizer with $2e-3$ learning rate for $4$ epochs. Our model had $98.40\%$ accuracy after the initial training. We pruned the model with $1\%$ tolerated accuracy loss. After pruning, the model was retrained with the same configuration. As a result, $92.04\%$ of weights were pruned, and the rest of them were quantized, the final accuracy was $97.37\%$. In Table~\ref{table:generalComparisonLenet}, it can be seen that we achieved a significant improvement in terms of memory usage and inference speed. After pruning and quantization, executable file size decreased by $12.45\times$. Inference run-times were calculated by taking executables’ average execution time.

\begin{table}[b]
\centering
\def\arraystretch{1.5}
\caption{\label{table:generalComparisonLenet}Comparison of initial and compressed LeNet-5 models.}
\begin{tabular}{ |c|c|c|c|c|c| } 
\hline
Model & Accuracy & Sparsity & Inference Time & Executable Size & Compression Rate \\
\hline
LeNet-5 & $98.40\%$ & $-$ & $4.98$ ms & $265$ KB & $-$ \\
\hline
Pruned LeNet-5 & $97.45\%$ & $91.92\%$ & $2.31$ ms & $46$ KB & $5.74\times$ \\
\hline
Pruned and Quantized LeNet-5 & $97.37\%$ & $92.04\%$ & $1.94$ ms & $21$ KB & $12.45\times$ \\
\hline
\end{tabular}
\end{table}

\par\medskip
In Table~\ref{table:layerComparisonLenet}, we can see that all layers are pruned above $91\%$ in the pruning phase. After the pruning phase, we achieved $91.92\%$ sparsity. Quantization also improved our sparsity, because some weights are quantized to zero. After quantization, our final sparsity increased to $92.04\%$. Total weight size is $61{,}470 * 4$ B $\approx 245$ KB in the original model. After pruning, total weight size drops to $4{,}967 * (4$ B$ + 1$ B$) \approx 25$ KB which is $9.9\times$ compression. After quantization, total weight size becomes $4{,}891 * (1$ B$ + 1$ B$) \approx 10$ KB which is $25.1\times$ compression. However, executable files do not achieve the same compression rate since they also include biases, input data, and header files, which are necessary for calculations.

\begin{table}
\centering
\def\arraystretch{1.5}
\caption{\label{table:layerComparisonLenet}Layer-wise analysis of LeNet-5 compression.}
\begin{tabular}[ht]{ |c|c|c|c|c|c| } 
\hline
\multirow{2}{*}{Layer}& LeNet-5 & \multicolumn{2}{c|}{Pruned LeNet-5} & \multicolumn{2}{c|}{Pruned and Quantized LeNet-5}\\
\cline{2-6}
    & Total Weights & Nonzero Weights & Sparsity & Nonzero Weights & Sparsity\\
\hline
CONV$_1$ & $150$ & $13$ & $91.33\%$ & $13$ & $91.33\%$ \\
\hline
CONV$_2$ & $2{,}400$ & $193$ & $91.96\%$ & $191$ & $92.04\%$ \\
\hline
FC$_1$ & $48{,}000$ & $3{,}884$ & $91.91\%$ & $3{,}815$ & $92.05\%$ \\
\hline
FC$_2$ & $10{,}080$ & $809$ & $91.97\%$ & $804$ & $92.02\%$ \\
\hline
FC$_3$ & $840$ & $68$ & $91.90\%$ & $68$ & $91.90\%$ \\
\hline
Total & $61{,}470$ & $4,967$ & $91.92\%$ & $4{,}891$ & $92.04\%$ \\
\hline
\end{tabular}
\end{table}

\par\medskip
We also evaluated our improved layer algorithms to see how they affect inference speed. These
algorithms were an important part of our compression since we would have to use more computational power without them. In Table~\ref{table:algoComparison}, we can see that improving layer calculations according to sparse format is important in terms of the inference speed of the model. In this model, FC layer calculations affect the performance the most since most of the pruning is done on FC layers.

\begin{table}
\centering
\def\arraystretch{1.5}
\caption{\label{table:algoComparison}Evaluation of improved layer algorithms.}
\begin{tabular}[ht]{ |c|c|c|c|c|c|c|c|c|c| } 
\hline
Model & LeNet-5 & \multicolumn{4}{c|}{Pruned LeNet-5} & \multicolumn{4}{c|}{Pruned and Quantized LeNet-5}\\
\hline
Improved Layers & None & None & CONV & FC & Both & None & CONV & FC & Both\\
\hline
Inference Time (ms) & $4.98$ & $275.40$ & $222.42$ & $63.70$ & $2.31$ & $272.78$ & $211.46$ & $61.06$ & $1.94$\\
\hline
Inference Speed Rate & $-$ & $0.02\times$ & $0.02\times$ & $0.08\times$ & $2.16\times$ & $0.02\times$ & $0.02\times$ & $0.08\times$ & $2.57\times$\\
\hline
\end{tabular}
\end{table}

\par\medskip
Finally, we compared our theoretical calculations with the executable files generated by our generator. In Table~\ref{table:exeComparisonLenet}, it can be seen that most of our improvement is on the \textit{.text} section of the executables, where all weights are stored as read-only data. There is a difference with our theoretical calculations since \textit{.text} section also includes biases, instructions, and test inputs. Rest of the \textit{.text} section is $\sim10$ KB for the original model, $\sim11$ KB for the pruned model, and $\sim8$ KB for the quantized model. Quantized model has less difference since the input data is also quantized to 8-bit. Contrary to the significant improvement in the \textit{.text} section, \textit{.data} sections in compressed models increase slightly, because more variables are used while receiving weights from CSC arrays dynamically during CONV and FC layers’ operations. Dynamically received weight kernels are stored in a smaller array than full size weight arrays; hence, the pruned model has a larger \textit{.bss} section. Quantized model, on the other hand, has a much smaller \textit{.bss} section since ping-pong buffers~\cite{hasanunlu} are 8-bit signed integer arrays instead of 32-bit signed floating-point arrays.

\begin{table}
\centering
\def\arraystretch{1.5}
\caption{\label{table:exeComparisonLenet}Comparison of models’ generated executable files.}
\begin{tabular}{ |c|c|c|c|c|c|c| } 
\hline
Model & .text & .data & .bss & .dec & .hex & Compression Rate \\
\hline
LeNet-5 & $256{,}031$ & $784$ & $8{,}832$ & $265{,}647$ & 40daf & $-$ \\
\hline
Pruned LeNet-5 & $35{,}751$ & $920$ & $9{,}632$ & $46{,}303$ & b4df & $5.74\times$ \\
\hline
Pruned and Quantized LeNet-5 & $17{,}945$ & $928$ & $2{,}456$ & $21{,}329$ & 5351 & $12.45\times$ \\
\hline
\end{tabular}
\end{table}

\par\medskip
We also compared our improved generator on a different architecture trained on CIFAR-10 dataset~\cite{cifar}. Unlu has compared his method with CMSIS-NN~\cite{cmsis} and shown to have a more efficient method in terms of RAM usage. This network has the following architecture:

\begin{verbnobox}[\fontsize{7pt}{7pt}\selectfont]
(0): Conv2d(3, 32, kernel_size=(5, 5), stride=(1, 1), padding=(2, 2))
(1): ReLU()
(2): MaxPool2d(kernel_size=2, stride=2)
(3): Conv2d(32, 16, kernel_size=(5, 5), stride=(1, 1), padding=(2, 2))
(4): ReLU()
(5): MaxPool2d(kernel_size=2, stride=2)
(6): Conv2d(16, 32, kernel_size=(5, 5), stride=(1, 1), padding=(2, 2))
(7): ReLU()
(8): MaxPool2d(kernel_size=2, stride=2)
(9): Flatten()
(10): Linear(in_features=512, out_features=10, bias=True)
\end{verbnobox}

\par\medskip
After running our generator on this model, we found out that the architecture is quite sensitive to the input data. Quantization of input activations halves the accuracy of the model; therefore, we did not quantize input activations like output activations. In Table~\ref{table:exeComparisonTest}, the comparison of test networks trained on CIFAR-10 is shown. Despite the fact that differences in all sections are similar to LeNet-5 compression, compression rate is smaller since the test network is more sensitive to pruning and input data, which is $\sim12$ KB, was not quantized. We achieved $88.33\%$ sparsity after pruning and quantization.

\begin{table}
\centering
\def\arraystretch{1.5}
\caption{\label{table:exeComparisonTest}Comparison of test network’s generated executable files.}
\begin{tabular}{ |c|c|c|c|c|c|c| } 
\hline
Model & .text & .data & .bss & .dec & .hex & Compression Rate \\
\hline
Original & $150{,}163$ & $760$ & $45{,}088$ & $196{,}011$ & 2fdab & $-$ \\
\hline
Pruned & $37{,}763$ & $864$ & $53{,}280$ & $91{,}907$ & 16703 & $2.13\times$ \\
\hline
Pruned and Quantized & $26{,}457$ & $872$ & $13{,}344$ & $40{,}673$ & 9ee1 & $4.82\times$ \\
\hline
\end{tabular}
\end{table}

\par\medskip
\cite{hasanunlu} also quantized test network’s weights when comparing his method with CMSIS-NN since Lai et al.~\cite{cmsis} quantizes weights of the test network. \cite{hasanunlu}’s weights utilization in ROM is $36$ KB and RAM utilization is $11.2$ KB. Our weights utilization in ROM is $20$ KB and RAM utilization is $14$ KB. Although our ROM utilization is $44\%$ less, RAM utilization is $25\%$ more since we need a small 8-bit array to receive weight kernels dynamically in CONV calculations. However, our RAM utilization is still $68\%$ less than CMSIS-NN. Unlu's total memory usage is $\sim52$ KB and our total memory usage is $\sim41$ KB ($23\%$ less). In our approach, there is a trade-off. In order to minimize ROM usage and maximize inference speed, a small amount of RAM must be sacrificed. Since our approach minimizes ROM usage by saving weights as CSC arrays, Algorithm~\ref{alg:conv} needs to load weights dynamically in RAM so that they can be used in further convolutional calculations without spending more memory and time.

\par\medskip
Additionally, our approach achieved a significant speed up while running on the on the same computer as LeNet-5. Original model executes in $57.09$ ms. Pruned model takes $15.72$ ms ($3.63\times$) and quantized model takes $12.99$ ms ($4.39\times$). This shows that our algorithms boosted the network more than LeNet-5. That is because this model has more and larger CONV layers than LeNet-5 which takes more time to calculate. From this result, we can deduce that speeding CONV layers up is more crucial than FC layers. 

\section{Conclusion}
Our work shows that applying compression algorithms on PyTorch models and efficiently converting them to C executable files makes their deployment on MCUs easier. First, we prune weights with the optimal sparsity percentage found by the binary search algorithm. Secondly, we quantize pruned weights and activations to 8-bit signed integers. Then, we save nonzero weight values in CSC format. Finally, we improve~\cite{hasanunlu}'s existing CONV and FC layer algorithms to use CSC arrays directly. As a result, we both decrease the memory usage by $12.45\times$ and increase the inference speed by $2.57\times$ without losing significant accuracy.

{\small
\bibliographystyle{unsrt}
\bibliography{egpaper_final}
}

\end{document}